\documentclass[letterpaper, 10 pt, conference]{ieeeconf}  

\IEEEoverridecommandlockouts                              

\usepackage{times}
\usepackage{multicol}
\usepackage[bookmarks=true]{hyperref}
\usepackage{graphics} 
\usepackage{amsmath} 
\usepackage{amssymb}  
\usepackage{mathtools}
\usepackage{float}
\usepackage[noend]{algpseudocode}
\usepackage{algorithm, algorithmicx}
\usepackage[font=small,labelfont=bf]{caption}
\usepackage{graphicx}
\usepackage{multirow}
\usepackage{bm}
\usepackage{booktabs}
\usepackage{array}
\usepackage{threeparttable}
\usepackage[dvipsnames]{xcolor}
\usepackage{cite}
\usepackage{subfig}

\definecolor{red}{rgb}{1, 0, 0}
\definecolor{green}{rgb}{0, 1, 0}
\definecolor{grey}{rgb}{0.8,0.8,0.8}
\definecolor{yellow}{rgb}{1,1,0}
\newcolumntype{C}[1]{>{\centering}m{#1}}
\algnewcommand\True{\textbf{true}\space}

\newcommand{\norm}[1]{\left\lVert{#1}\right\rVert}
\newcommand{\q}{\bm{q}}
\newcommand{\Q}{\bm{Q}}
\newcommand{\voxel}{\bm{v}}
\newcommand{\Vunknown}{\bm{V}_{\text{unknown}}}
\newcommand{\Vfree}{\bm{V}_{\text{free}}}
\newcommand{\Vocc}{\bm{V}_{\text{occ}}}
\newcommand{\Fugv}{\mathbb{F}_{\text{ugv}}}
\newcommand{\Fuav}[1][]{\mathbb{F}_{\text{uav}}^{#1}}
\newcommand{\Fagent}{\mathbb{F}_{\text{agent}}}
\newcommand{\Frontier}{\mathbb{F}}
\newcommand{\Cagent}{\mathbb{C}_{\text{agent}}}
\newcommand{\Cuav}{\mathbb{C}_{\text{uav}}}
\newcommand{\Cugv}{\mathbb{C}_{\text{ugv}}}
\newcommand{\Iv}{\mathbb{I}_{\bm{v}}}
\newcommand{\pC}{\bar{\bm{p}}_C}
\newcolumntype{P}[1]{>{\centering\arraybackslash}p{#1}}

\title{\LARGE \bf  
Coordinated Aerial-Ground Robot Exploration via Monte-Carlo View Quality Rendering
}

\author{Di Deng, Zhefan Xu, Wenbo Zhao, and Kenji Shimada
\thanks{Department of Mechanical Engineering, Carnegie Mellon University, 5000 Forbes Ave, Pittsburgh, PA 15213, USA.,
{\tt\small dengd@andrew.cmu.edu}. 
}
\thanks{The authors would like to thank TOPRISE Co., LTD. for their partial financial support for this work.}
}

\begin{document}
\maketitle
\thispagestyle{empty}
\pagestyle{empty}

\noindent \begin{abstract}
We present a framework for a ground-aerial robotic team to explore large, unstructured, and unknown environments. In such exploration problems, the effectiveness of existing exploration-boosting heuristics often scales poorly with the environments' size and complexity. This work proposes a novel framework combining incremental frontier distribution, goal selection with Monte-Carlo view quality rendering, and an automatic-differentiable information gain measure to improve exploration efficiency. Simulated with multiple complex environments, we demonstrate that the proposed method effectively utilizes collaborative aerial and ground robots, consistently guides agents to informative viewpoints, improves exploration paths' information gain, and reduces planning time.

\end{abstract}

\section{Introduction}
\begin{figure}
\centering
\vspace{-0.2cm}
\subfloat[Storage house scene]{
\includegraphics[width=0.45\linewidth]{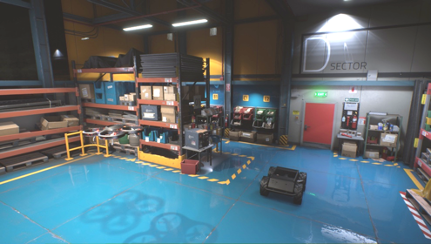}
}   
\subfloat[Geometry overview] {
	\includegraphics[width=0.45\linewidth]{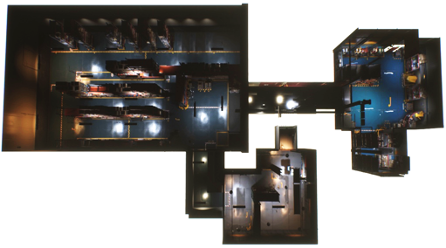}
}\\
\vspace{-0.1cm}
\subfloat[Exploration paths of the UAV (red) and the UGV (white)] {
	\includegraphics[width=0.45\linewidth]{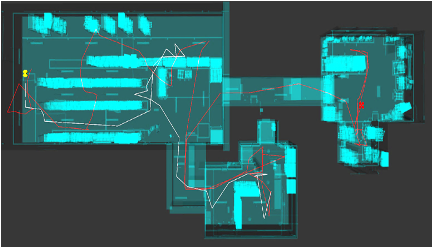}
}
\subfloat[Reconstructed Environment] {
	\includegraphics[width=0.45\linewidth]{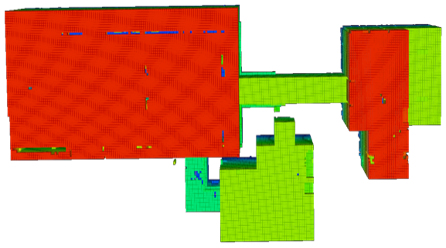}
}
\caption{Path planning for storage house exploration with a RGB-D camera-equipped quadrotor and a lidar-equipped ground robot.}
\label{fig:storage}
\vspace{-0.6cm}
\end{figure}

\noindent The rapid progress in sensing technologies and affordable onboard sensors has encouraged an extensive adoption of heterogeneous-agent cooperation for coverage planning problems \cite{palaciosmulti} for unstructured environments. The environment is assumed to be unknown because the environment's prior knowledge is inaccurate due to unknown obstacles or uncertainty in sensor measurements. Coverage planning for exploration can be applied to tasks such as infrastructure inspection \cite{thrun2004autonomous}, seabed coverage \cite{ferri2017cooperative}, and disaster survivor search and rescue \cite{liu2013robotic}. These tasks require teams of heterogeneous robots equipped with depth sensors to explore an unstructured 3D environment efficiently. Fig. \ref{fig:storage} shows an example of a target environment and optimized exploration paths generated by our method.

Compared with homogeneous robotic systems, heterogeneous agents can accomplish complex tasks more efficiently because different agents can complement each other \cite{rizk2019cooperative}. For instance, Unmanned Aerial Vehicles (UAVs) can navigate complex three-dimensional environments but are limited by their minimum hovering height. On the other hand, Unmanned Ground Vehicles (UGVs) can explore regions close to the floor but have restricted views \cite{tokekar2016sensor}. Autonomous Underwater Vehicles (AUVs) can take a closer look at the sub-sea world but suffer from limited localization accuracy and operating speed \cite{ropero2019terra}. 

Exploration of unknown environments by robotic agents typically involves planning trajectories for agents using the current partially-known map. 
Maximizing the exploration of unknown regions hinges on the prudent selection of goals for these trajectories.
As rigorously finding the global maximum of a function which quantifies information gain is difficult, existing goal selection methods often rely on heuristics. 

A common heuristic is to navigate agents directly to the nearest \textit{frontier}, defined as the boundary between known and unknown regions of the current map \cite{yamauchi1997frontier}. However, the nearest frontier is not guaranteed to be accessible to the robot. Even if it is, it may not be as informative as other viewpoints. Another common heuristic is to sample many feasible viewpoints within the known free region of the map, and evaluate for all samples their information gain, which is defined as the number of visible frontiers from a viewpoint. The sample with the maximum information gain is chosen as the goal \cite{bircher2016receding, witting2018history, selin2019efficient, schmid2020efficient}. However, as samples are biased towards explored free regions, sample efficiency tends to degrade as exploration progresses.

\textbf{Contribution}. 
We propose a real-time frontier distribution and goal selection strategy with Monte-Carlo view quality rendering, which scales well with the environment's dimension and complexity, and consistently guides heterogeneous agents to informative viewpoints in large cluttered environments. Selecting exploration goals with Monte-Carlo view quality rendering overcomes a limitation of conventional frontier-based method, uninformative, and inaccessible goal generation. Besides, it avoids inefficient sampling and repetitive computation of sampling-based methods. The paths from the robots' current configurations to the selected goals are further optimized with a frontier-based automatic-differentiable information gain measure \cite{deng2020robotic}. Compared with paths generated using a single agent, paths generated and optimized with the proposed method are consistently superior in planning time and coverage efficiency.


\section{Related Work}
\noindent Task distribution is one of the main concerns for multi-robot exploration problems. \cite{yamauchi1998frontier} navigates agents towards their nearest frontier independently. Even though conflicts among robots' paths are resolved with reactive collision avoidance, the lack of cooperation often leads to repetitive explorations of the same region. 
In \cite{burgard2000collaborative}, a utility function is used to rank task allocation to individual agents. The utility function is initialized to the information gain of a viewpoint and is discounted as the region is assigned to other agents. \cite{solanas2004coordinated} utilizes clustering to distribute target exploration regions. \cite{zlot2002multi} and \cite{simmons2000coordination} distribute the utility computation to individual robots and use the estimated utility as ``bids." A global executive receives all bids and makes global decisions to maximize the total utility while minimizing interference with one another. However, these approaches do not estimate or optimize the information gain along paths.


The cooperation of UAVs and UGVs can vastly improve the exploration efficiency and mapping accuracy by taking advantage of different motion and sensing abilities of heterogeneous agents. Research work on the cooperation of heterogeneous agents includes hierarchical path planning. For example, Ghamry \textit{et al}. \cite{ghamry2016cooperative} use UAVs and UGVs to detect and fight forest fires. UAVs uniformly scan an enclosed target region, while UGVs recognize fire regions based on the aerial footage and plan trajectories for UAVs to put out the fire. Sujit \textit{et al}. explore the ocean using a UAV and AUVs \cite{sujit2009uav}. The UAV skims through a large region quickly and identifies particular regions of interest for AUVs to survey closely. Qin \textit{et al}. first utilize a UGV to obtain a coarser map and then deploy a UAV to refine the map \cite{qin2019autonomous}. Their approaches improve the map's quality but suffer from a long exploration time.

Instead of the sequential deployment of heterogeneous agents, many researchers propose simultaneous exploration. Butzke \textit{et al}. \cite{butzkey20153} conduct 3D exploration in an unknown environment with an aerial-ground robot system using a frontier-based \cite{yamauchi1997frontier} goal identifier and a lattice-based AD$^*$ algorithm. Grocholsky \textit{et al}. \cite{grocholsky2006cooperative} search and localize targets in a given area with a decentralized control of a network of UAVs and UGVs with different navigation speeds and target measurement accuracies. By sharing a probabilistic map of the target and driving the vehicles by combined mutual information gain gradients, the sensing network can detect targets faster than a single UGV or a single UAV. \cite {li2016hybrid} applies a genetic algorithm to the path planning for a global aerial-ground robot team and optimizes the local rolling path.

\section{Overview of Proposed Planner}
\noindent The proposed planning algorithm uses an aerial-ground robot system consisting of one ground agent (UGV) and one aerial agent (UAV). Both agents are equipped with a 3D range sensor, a lidar on the UGV and an RGB-Depth camera on the UAV. Their goal is to explore an unknown bounded 3D space, $\bm{V} \subset \mathbb{R}^3$, determining the subsets of $\bm{V}$ that are free ($\Vfree \subseteq \bm{V}$), occupied ($\bm{V}_{\text{occ}} \subseteq \bm{V}$) and unknown ($\Vunknown \subseteq \bm{V}$). The space, $\bm{V}$, consists of voxels, $\bm{v} \in \bm{V}$, with minimum edge length $\varrho$, defined as the resolution of the space. The occupancy status of $\bm{V}$ is represented with a probabilistic occupancy map denoted by $\mathbb{M}$ \cite{thrun2002probabilistic}. 

\begin{figure}[tp!]
    \centering
    \subfloat[] {
	\includegraphics[width=0.45\linewidth]{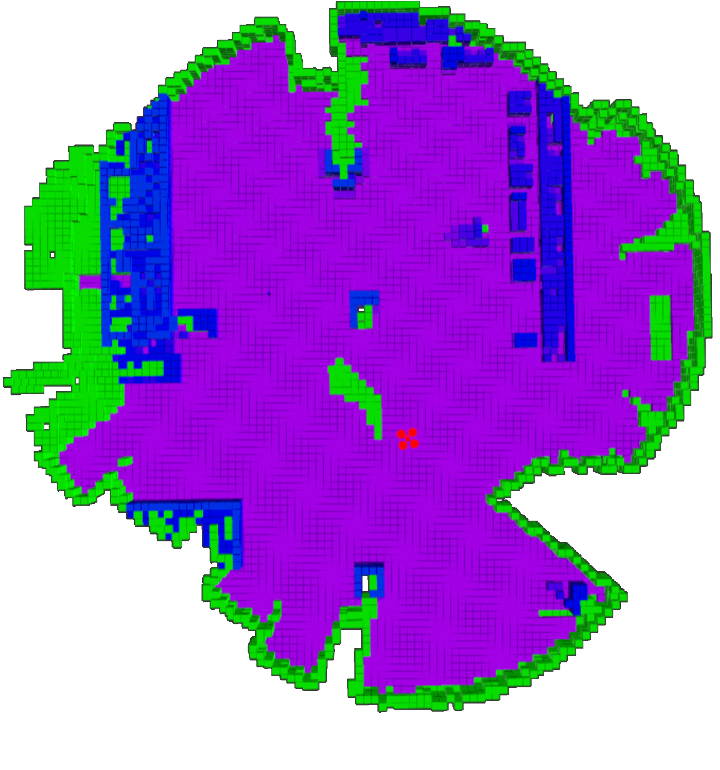}
    }
    \subfloat[] {
	\includegraphics[width=0.45\linewidth]{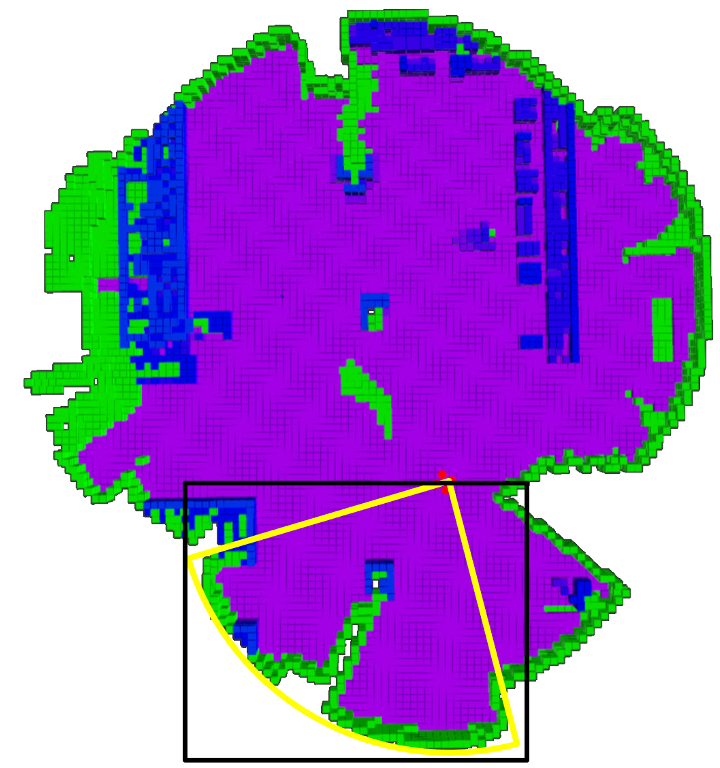}
    }
    \caption{Frontiers and incremental frontier update. 
    Purple and blue voxels are explored occupied voxels. Green voxels are the frontiers. (\textbf{a}) Original occupancy map and frontier voxels.  (\textbf{b}) Updated occupancy map and frontier voxels. The camera's view frustum is shown as the yellow circular sector. 
    }
    \label{fig:incremental_frontier}
\end{figure}

A voxel, $\voxel$, is called a \textit{frontier} if $\voxel \in \Vunknown$ and there exists $\voxel' \in \Vfree$ such that $\voxel$ and $\voxel'$ are adjacent. Frontiers represent the boundary between explored free space $\Vfree$ and unexplored space $\Vunknown$, as illustrated in Fig. \ref{fig:incremental_frontier}(a). The set of frontiers is denoted by $\mathbb{F}$.

An overview of the collaborative aerial-ground robot exploration framework is presented in Alg. \ref{algo:gradient_planning}, which requires the starting configurations of the agents, $\q_{0} = [\q_{\text{uav(0)}}, \q_{\text{ugv(0)}}]$, as inputs. 
We will also refer to configurations of agents as \textit{viewpoints}. A \textit{path}, $\Q$, is an ordered set of viewpoints: $\Q = \{\q_1, \dots, \q_n\}$. The \textit{information gain} of a path is defined as the sum of the number of visible frontiers from all viewpoints in the path.

\begin{algorithm}
\caption{Gradient-Based Space Coverage}
\label{algo:gradient_planning}
\begin{algorithmic}[1]
\Require $\q_0$ 
\Ensure $\mathbb{M}$
\State $\mathbb{M}$ = Octomap($\varrho$), IG = $\inf$
\State $\mathbb{M}\gets$ UpdateMap($\q_0, \mathbb{M}$)
\While{IG $> \epsilon$}
\State $\mathbb{F}_{\text{ugv}}, \mathbb{F}_{\text{uav}} \gets$ DistributeFrontier($\mathbb{M}, \q_0$) 
\State $\q_{\text{goal}} \gets$ SelectGoal($\q_0, \mathbb{F}_{\text{ugv}}, \mathbb{F}_{\text{uav}}, \mathbb{M}$)
\State $\Q \gets$ GlobalPlanner($\q_0$, $\q_{\text{goal}}$, $\mathbb{M}$) \quad // Sec. \ref{subsec.global_path}
\State $\Q^*$, IG $\gets$ GradientPathOptimizer($\Q, \mathbb{M}$) \; // Sec. \ref{sec:path_optimization}
\For{$\q_i$ in $\Q^* = \{\q_1, \ldots \q_n\}$}
\State $\mathbb{M} \gets$ UpdateMap($\q_i$, $\mathbb{M}$)
\EndFor
\State $\q_0=\q_{\text{goal}}$
\EndWhile
\end{algorithmic}
\end{algorithm}

\begin{figure*}
\vspace{0.2cm}
\centering
\includegraphics[width = 0.8\linewidth]{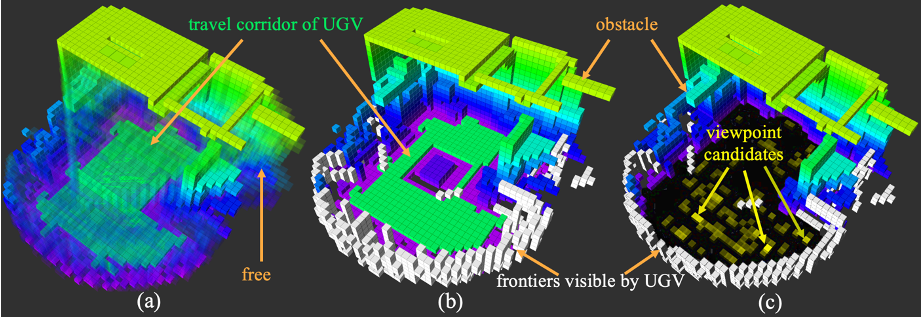}
\caption{UGV's next-best-view rendering. (\textbf{a}) UGV's travel corridor. Green voxels are the travel corridor of the UGV. They are explored free voxels, defining the collision-free regions for the UGV. Colored voxels are explored occupied voxels; transparent voxels are free voxels. (\textbf{b}) UGV's visible frontiers. Frontiers are denoted as white voxels, visible by the UGV that travels within the trajectory corridor. (\textbf{c}) Viewpoints' information gain. The yellow voxels on the travel corridor are the viewpoints that can observe at least one frontier voxel. The lower opacity of the yellow voxel, the more frontiers it observes. }
\label{fig:trajectory_corridor}
\end{figure*}

Before exploration starts, occupancy map $\mathbb{M}$ with resolution $\varrho$ is initialized with the range measurements of the agents at $\q_{0}$ (Lines 1-2).
In every exploration iteration, frontiers $\mathbb{F}$ are first distributed to the UGV and the UAV, $\Fuav$ and $\Fugv$ (Line 4), which is discussed in details in Sec. \ref{sec:frontier_distribution}.
Next, both agents' goals, $\q_{\text{goal}} = [\q_{\text{uav(goal)}}, \q_{\text{ugv(goal)}}]$, are selected using the proposed Monte-Carlo view quality renderer (Line 5), which is presented in Sec. \ref{sec.render_nbv}. 
Then, a sampling-based algorithm like RRT plans collision-free paths, $\Q$, from $\q_{0}$ to  $\q_{\text{goal}}$ (Line 6), which is further optimized with respect to a frontier-based differentiable information gain measure (Line 7) \cite{deng2020robotic}. 
Finally, the optimized path, $\Q^*$, is executed by the agents to update the occupancy map $\mathbb{M}$ (Lines 8-9). As shown in Fig. \ref{fig:incremental_frontier}(b), the set of frontiers $\Frontier$ is incrementally updated with new range measurements from agents: newly detected frontiers are added, while voxels that are no longer frontiers are removed. 
Our implementation uses the Octomap library \cite{hornung2013octomap}, which stores occupancy map in an octree, a spatial data structure efficient for queries and updates. 
This process repeats until the path's information gain falls below a threshold, $\epsilon$ (Line 3).

\section{Frontier Distribution} \label{sec:frontier_distribution}
\noindent Frontiers in $\Frontier$ are distributed to both robotic agents in order to guide the planning of their exploration path. Frontier assignment to UGVs is prioritized, as UGVs are constrained to moving on the ground, thereby more likely to have their views obstructed by obstacles. 

\begin{algorithm}
\caption{Frontier Distribution}
\label{algo:frontier_distribution}
\begin{algorithmic}[1]
\Require $\Frontier, \Cagent$ 
\Ensure $\Fagent$
\State $\Fagent \gets []$
\For{$\voxel$ \textbf{in} $\Frontier$}
    \State $\Iv \gets \text{FindIntersection}(\voxel, \Cagent)$
    \If{$\Iv = \emptyset$} 
        \State \textbf{continue}
    \EndIf
    \For{$\voxel'$ \textbf{in} $\Iv$}
        \If{RayCast($\voxel, \voxel'$)}
            \State $\Fagent$.append($\voxel$)
            \State \textbf{break}
        \EndIf
    \EndFor
\EndFor
\end{algorithmic}
\end{algorithm}

\begin{figure}
\centering
\subfloat[] {
\includegraphics[width=0.45\linewidth]{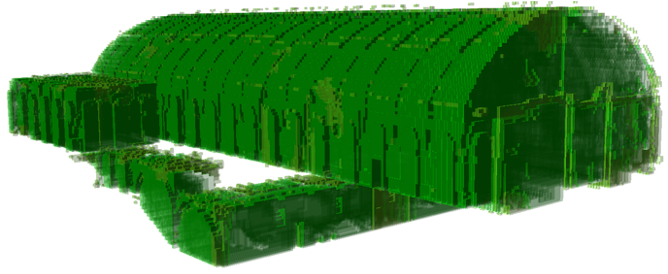}
}
\subfloat[] {
\includegraphics[width=0.45\linewidth]{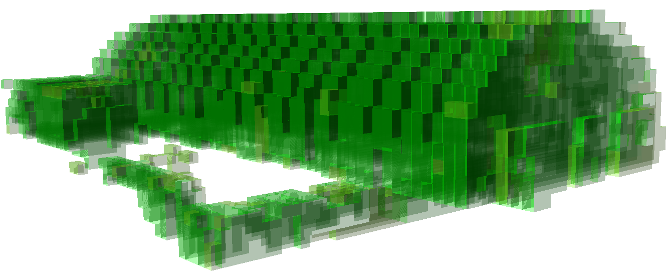}
}
\caption{UAV's travel corridor. (\textbf{a}) Original free voxels. (\textbf{b}) Free voxels after limiting depth cutoff.}
\label{fig:uav_corridor}
\end{figure}

The frontier distribution procedure for an agent is detailed in Alg. \ref{algo:frontier_distribution}, which requires as input the current frontiers, $\Frontier$, and the \textit{travel corridor} of the agent. The agent's travel corridor, denoted by $\Cagent \subset \Vfree$, is a subset of the known collision-free region reachable by the agent. For the ground agent, $\Cugv$ is always on a plane parallel to the ground. An example of $\Cugv$ is given in Fig. \ref{fig:trajectory_corridor}. For the aerial agent, $\Cuav$ consists of free voxels with edge length on par with the size of the UAV's collision geometry. It can be obtained from Octomap's multi-resolution queries, limiting the cutoff depth to traverse through a coarser map, as shown in Fig. \ref{fig:uav_corridor}. 

\begin{figure}
\centering
\vspace{0.2cm}
\subfloat[UGV. $h$ is the height (z-coordinate) of $\voxel$. $h_0$ is the height of the UGV's range sensor, which is constant as the UGV is constrained to move on the ground.]{
	\includegraphics[width=0.9\linewidth]{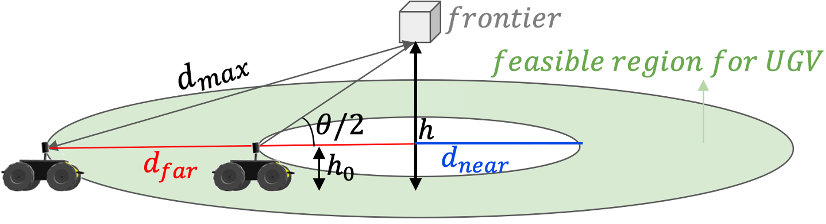}
	\label{fig:feasible_region:ugv}
}\\
\subfloat[UAV.]{
	\includegraphics[width=0.7\linewidth]{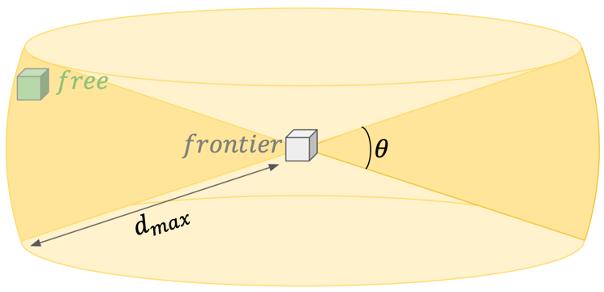}
	\label{fig:feasible_region:uav}
}
\caption{Feasible region of Frontier $\voxel$ with respect to an agent. $d_{\text{max}}$ is the maximum range of the agent's onboard sensor, and $\theta$ the sensor's vertical field of view.}
\label{fig:feasible_region}
\vspace{-0.4cm}
\end{figure}

For each $\voxel \in \Frontier$, there exists a feasible region where $\voxel$ is visible to the agent, which is shown in Fig. \ref{fig:feasible_region}. This region only considers the geometry of the view frustum of the agent's range sensor without considering occlusion from obstacles in $\Vocc$ yet. Line 3 returns $\Iv$, the intersection of the voxel's feasible region and $\Cagent$. 

If $\Iv$ is empty (Line 4), there exists no viewpoint that is both visible to $\voxel$ and reachable by the agent. On the other hand, if $\Iv$ is non-empty, we further check for occlusion by obstacles in $\Vocc$ in Line 7. The function RayCast returns true if the line segment connecting $\voxel$ and $\voxel'$ does not intersect with any element of $\Vocc$. A frontier voxel that passes this test is added to $\Fagent$ (Line 8).

\section{UGV Goal Selection with Monte-Carlo View Quality Rendering} \label{sec.render_nbv}
\noindent Inspired by Monte-Carlo ray tracing techniques commonly used in computer graphics to generate photo-realistic image \cite{liu2019cinematic}, we propose a new technique to compute the information gain of voxels in the UGV's travel corridor, $\Cugv$. As $\Cugv$ lies on a plane, voxels in $\Cugv$ can be treated as pixels in an image. Analogously, voxels in $\Fugv$ are treated as point light sources, and voxels in $\Vocc$ and $\Vunknown$ as obstacles capable of blocking light. Accordingly, the more light $\voxel \in \Cuav$ receives, the higher the achievable information gain when the UGV is at $\voxel$. 

Specifically, Monte-Carlo view quality rendering consists of the following steps:
\begin{enumerate}
\item For each $\voxel \in \Cuav$, $\Fugv(\voxel)$, a subset of $\Fugv$ with $n_r$ elements, is formed from $\Fugv$ using rejection sampling: for each randomly chosen $\voxel' \in \Fugv$, $\voxel'$ is added to $\Fugv(\voxel)$ if it would be inside the feasible region of the UGV's sensor placed at $\voxel$. Here $n_r$ represents the number of ray casts per voxel.

\item A ray is then cast from $\voxel$ to each  $\voxel' \in \Fugv(\voxel)$, and the information gain (which is initialized to 0) of $\voxel$ is incremented by 1 if the ray from $\voxel$ to $\voxel'$ does not hit any obstacles.
\end{enumerate}

\begin{figure}[h]
\vspace{0.2cm}
    \centering
    \includegraphics[width=\linewidth]{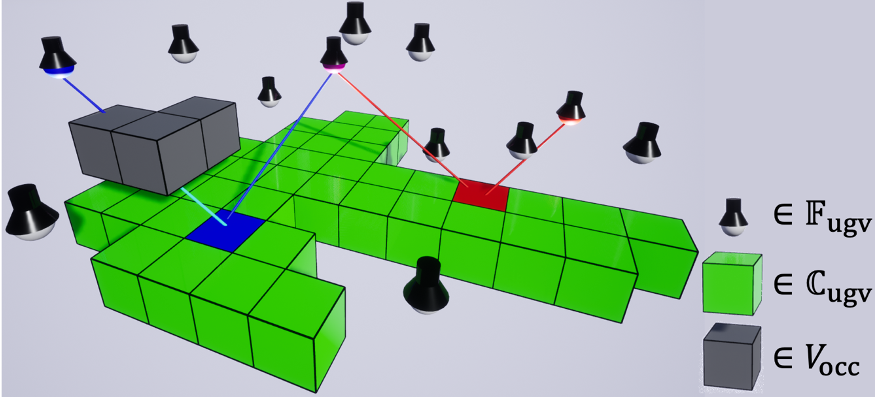}
    \caption{Monte-Carlo view quality rendering with $n_r=2$. The rendering of the blue and red voxels are highlighted. For each voxel, two light sources are randomly chosen, which are shown as lit up. Both light sources in $\Fugv(\voxel_\text{red})$ are visible from $\voxel_\text{red}$, giving it an information gain of 2. In contrast, only one light source in $\Fugv(\voxel_\text{blue})$ is visible from $\voxel_\text{blue}$, leaving $\voxel_\text{blue}$ with an information gain of 1.}
    \label{fig:MC_raycast}
\end{figure}

As shown in Fig. \ref{fig:MC_raycast}, this process resembles the Monte-Carlo integration that calculates the radiance of pixels in an image. An example of a view-quality image generated by the proposed rendering procedure is shown in Fig. \ref{fig:trajectory_corridor}(c), where brightness of the voxels is proportional to their information gain.

After calculating the information gain of every element in $\Cugv$ with Monte-Carlo rendering, the quality of all viewpoint candidates are evaluated using:
\begin{equation}
\label{eqn.IG}
    \text{ViewQuality}(\q) =  e^{-\lambda\norm{{\q-\q_0}}}\text{IG}(\q),
\end{equation}
where $\text{IG}(\q)$ is the information gain of the UGV at viewpoint $q$, and the exponential term penalizes the distance between the viewpoint, $\q$, and the UGV's current configuration, $\q_\text{UGV(0)}$. The view quality measure is the number of visible frontier voxels discounted by the distance to $\q_\text{UGV(0)}$. The UGV goal, $\q_\text{goal}$, is selected as the viewpoint with the highest view quality.

Our current implementation utilizes Octomap's CPU-based ray casting routines. As hardware-accelerated ray casting has already been supported on standard Desktop GPUs, we believe the proposed rendering method would run significantly faster and with much more ray casts per voxel. 

\section{Frontier-based UAV goal selection}
\noindent The rendering procedure presented in Section \ref{sec.render_nbv} computes view quality for the UGV and can be applied in principle to the UAV as well. However, as the UAV's travel corridor, $\Cuav$, is a 3D volume instead of a 2D plane, rendering view quality for every voxel in $\Cuav$ would be too time-consuming using a CPU-based ray-casting subroutine. We therefore employ a more conventional approach to select UAV exploration goals based on frontier clusters. This drives the UAV to the cluster with the highest frontier voxel density. 

\begin{figure}[h]
\centering
\includegraphics[width = \linewidth]{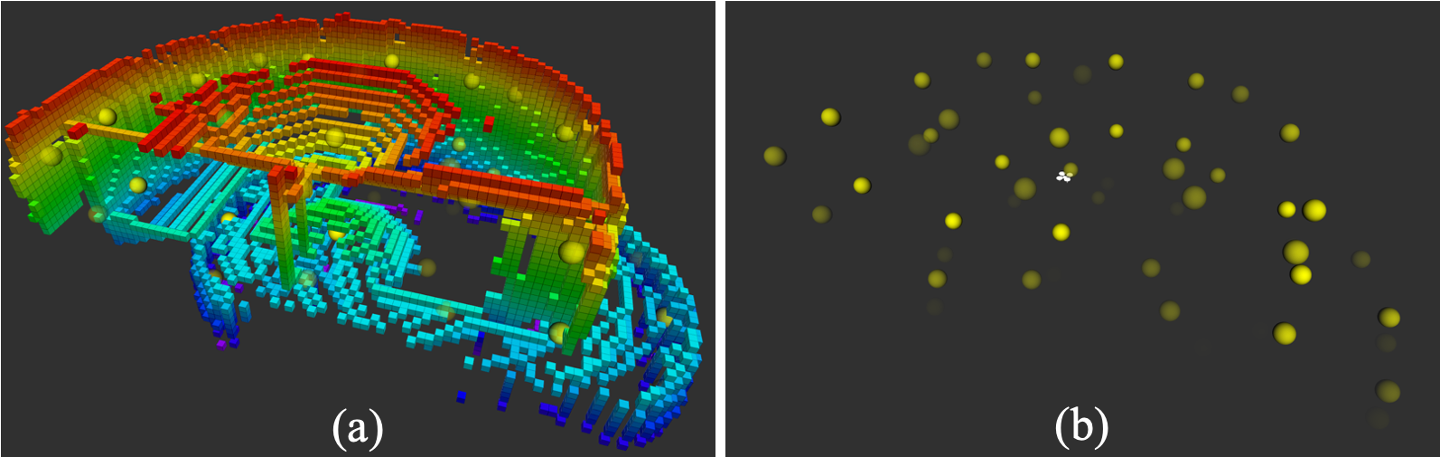}
\caption{UAV frontier clusters. (\textbf{a}) The frontier cluster centers, $\pC^i$, are shown as yellow spheres. Frontiers are shown as transparent colored voxels. (\textbf{b}) The centers of frontier clusters. The opacity of spheres is proportional to the number of frontiers within that cluster.  The less transparent the sphere, the denser the frontier voxels within the cluster.}
\label{fig:light_driven_uav}
\end{figure}

A frontier cluster, $\Fuav[i]$, is defined as frontiers within the same leaf node of a coarser subdivision of the frontier map by cutting the octree at a higher level. Let $\pC^i$ denote the geometric center of $\Fuav[i]$ and $n_i$ the number of frontiers in $\Fuav[i]$. The view quality of cluster centers can then be evaluated using (\ref{eqn.IG}) with $f(\bm{p})=n_i$. An example of frontier clusters is shown in Fig. \ref{fig:light_driven_uav}. 

However, cluster centers sometimes are non-free voxels unsuitable to be the goal of the UAV. In addition, even if the viewpoint is a free voxel, it might be occluded from frontiers within the cluster. Thus, we search for the frontier closest to $\pC^{*}$, the center of the cluster with the highest view quality:
\begin{equation}
    \voxel^* = \underset{\voxel \in \Fuav[*]}{\text{argmin.}} \; \norm{\voxel - \pC^{*}},
\end{equation}
and try to find $\voxel \in \Cuav$ from which $\voxel^*$ is visible to the UAV. If found, then $\voxel$ is chosen as $\q_{\text{goal}}$. If not, we try to find $\voxel \in \Cuav$ from which the frontier second most closest to $\pC^*$ is visible. This process is repeated until a $\voxel' \in \Fuav[*]$ visible from $\Cuav$ is found.




\section{Gradient-based Path Optimization} \label{sec:path_optimization}
\subsection{Global-Path Generation} \label{subsec.global_path}
\noindent The path from the initial point, $\q_0$, to the goal point, $\q_\text{goal}$, is generated using RRT. The original path, $\Q = \{\q_0, \dots, \q_\text{k}\}$, for both the UGV and UAV are planned with the Open source Motion Planning Library (OMPL) \cite{sucan2012open}, with the valid states defined as all voxels within the collision geometry are free voxels from $\textbf{V}_{\text{free}}$.
\subsection{Frontier-based Viewpoint's Orientation Optimization}
\noindent Our gradient path optimization algorithm adopts the frontier-based automatic-differentiable information gain measure \cite{deng2020robotic} for exploring an unknown 3D environment and increasing visible frontiers along the path. The automatic-differentiable information gain of a viewpoint is achieved by adding a fuzzy logic filter to count visible frontier voxels surrounding a viewpoint so that its gradient is efficiently computed. As only the orientations of paths are optimized, and the onboard sensor of the UGV is 360$^{\circ}$, only the UAV's paths are optimized. 

Since the starting point, $\q_0$, and the goal point, $\q_k$, are fixed, the optimization program's decision variables are defined as $\Q = \{\q_1, \ldots, \q_{k-1}\}$. The optimization problem is thus defined as:
\begin{equation}
\label{eqn:optimization}
    \underset{\Q}{\text{minimize}} \; -\text{IG}_{\text{path}}(\Q),
\end{equation}
so that the information gain of a planned global path is optimized.
The optimization problem formulated in Eqn. ($\ref{eqn:optimization}$) is initialized with $\{\q_1, \dots, \q_{k-1}\}$ and solved using non-linear solvers such as IPOPT \cite{wachter2006implementation}.


\section{Results and Discussion}
\noindent This section evaluates the performance of the proposed path optimization method in multiple large complex 3D simulation environments.

\subsection{Simulation design and implementation details}
\noindent In our simulation, the UAV has four degrees of freedom: $[x, y, z, \theta]$, and the UGV has three degrees of freedom : $[x, y, 0, \theta]$. The UAV is equipped with a depth camera with a maximum depth range of 10m and a field of view of $[\frac{\pi}{2},\frac{2\pi}{5}]$ in XZ and YZ planes, respectively. The UGV is equipped with a 360$^{\circ}$ 3D lidar with a maximum range of 6m and a vertical field of view of [-20$^{\circ}$, 20$^{\circ}$]. The UAV's collision geometry is a cube with an edge length of 0.8m, while the collision geometry of the UGV is 1.0m$\times$1.0m$\times$0.7m. The resolution of the map, $\mathbb{M}$, is $\varrho = 0.3\text{m}$. 

\begin{figure}[htp!]
\vspace{0.2cm}
    \centering
    \includegraphics[width=\linewidth]{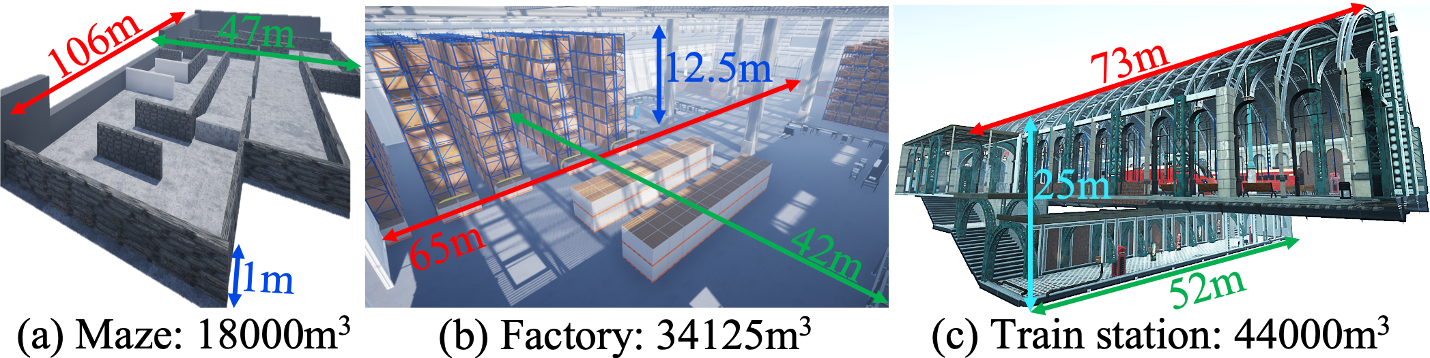}
    \caption{Scenery views}
    \label{fig:scene}
    \vspace{-0.4cm}
\end{figure}

Fig. \ref{fig:scene} illustrates the overall geometries and dimensions of three test environments: maze, factory, and train station. 
The train station is a cluttered environment with different levels. Our algorithm is simulated using Unreal Engine 4.24 \cite{karis2013real} and Airsim \cite{shah2018airsim}.

\subsection{Path Optimization}
 \noindent The frontier-based automatic-differentiable path optimization algorithm significantly improves the global path's information gain generated using RRT, as demonstrated in Table. 
\ref{tab:optimize_compare}. Collaborative exploration improves the UAV's information gain by 10.5\%, 30.0\%, and 34.5\% to cover 64\%, 93\%, and 81\% of the maze, the factory, and the train station. Compared with the global paths' lengths covering 57\%, 72\%, and 66\% of the three environments, the optimized paths' lengths are reduced by 12\%, 70\%, and 56\%.

\begin{table}[htp!]
\renewcommand\arraystretch{1.2} 
\begin{center}
\caption{Comparison of the UAV's information gain and path length improvement ratio before and after the proposed optimization during collaborative exploration.} \label{optimization_comparison_table}
\begin{tabular}{ |P{1.2cm} P{1cm}|P{1cm} P{1cm} P{1.5cm}|}
 \hline
  & & Maze & Factory & Train station\\
 \hline
 \multirow{2}{2em}{Information Gain} & before & 57\%  & 72\% & 66\% \\
 & after & \textbf{64}\% & \textbf{93}\% & \textbf{81}\% \\
 \hline
 \multirow{2}{2em}{Path length} & before &   286m& 777m & 713m\\
 & after &  \textbf{252}m & \textbf{232}m & \textbf{316}m    \\
 \hline
\end{tabular}
\label{tab:optimize_compare}
\end{center}
\vspace{-0.2cm}
\end{table}


\subsection{Exploration}
\begin{figure}[htp!]
    \centering
    \includegraphics[width=\linewidth]{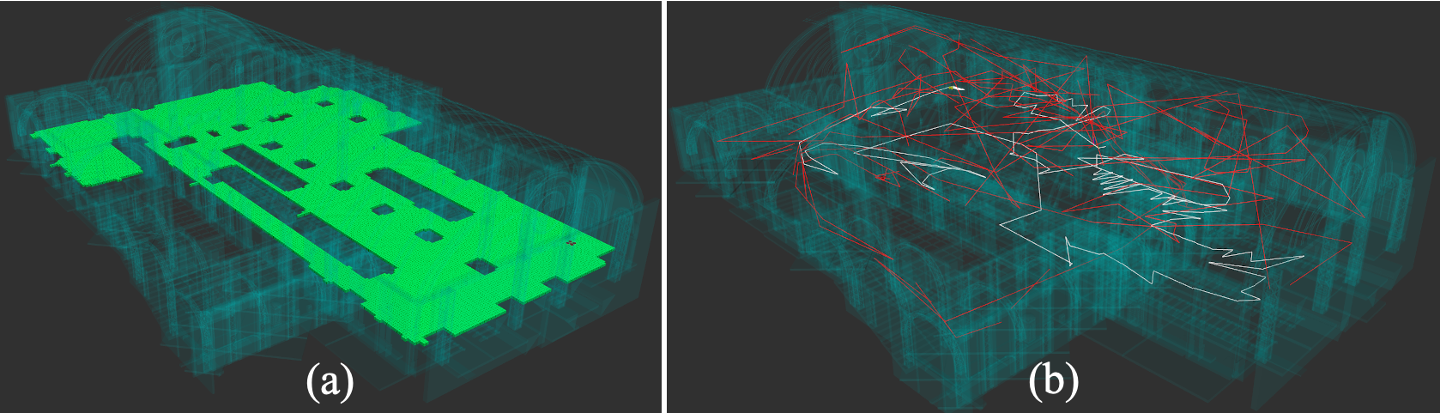}
    \caption{Exploration paths of the UGV-UAV system in the train station. (\textbf{a}) Travel corridor of the UGV. (\textbf{b}) Paths of the UAV (red lines) and UGV (white lines).}
    \label{fig:path}
    \vspace{-0.4cm}
\end{figure}

\noindent 
The exploration efficiency is vastly increased using collaborative exploration. The UAV's exploration path length is reduced from 1,147m using a UAV alone to 663m when the UAV works together with a UGV. As shown in Fig.\ref{fig:path}(b), the UAV's optimized collision-free path is the red line, and the UGV's path on the travel corridor, green voxels in Fig. \ref{fig:path}(a), are the white lines. 

\begin{figure*}
\vspace{0.2cm}
    \centering
    \includegraphics[width=\textwidth]{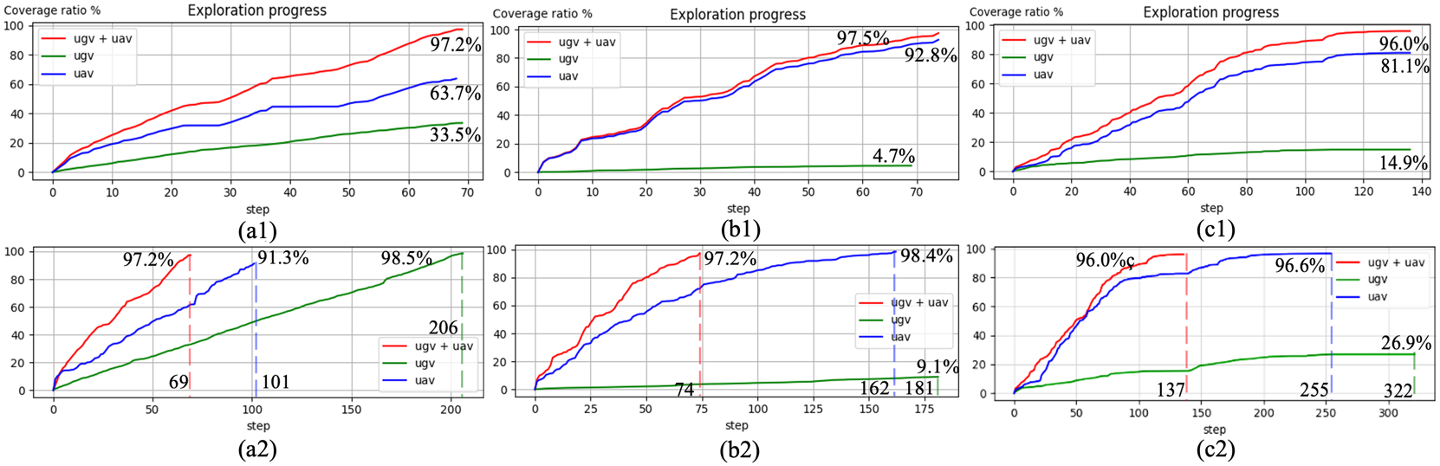}
    \caption{Comparison of the exploration performance in (\textbf{a}) maze, (\textbf{b}) factory, and (\textbf{c}) train station. (\textbf{a1}), (\textbf{b1}), and (\textbf{c1}) are the contributions of the UAV and UGV during collaboration exploration; (\textbf{a2}), (\textbf{b2}), and (\textbf{c2}) are comparisons of the exploration progress with a UAV alone, a UGV alone and a UGV-UAV team collaborative exploration in planning steps.}
    \label{fig:result}
    \vspace{-0.2cm}
\end{figure*}

 The collaborative robot system also significantly reduces path planning time by reducing planning steps. Each step consists of frontier distribution, next-best-view rendering, global path generation, and path optimization. As shown in Fig. \ref{fig:result}, we compare the UGV-UGV system, a single UGV, and a single UAV in terms of the exploration progress against planning iterations. As shown in Figs. \ref{fig:result}(a2), (b2), and (c2), the collaborative system has at least 30\% and can reach up to 54\% less path planning steps compared with the UAV alone. Moreover, comparing Fig. \ref{fig:result}(b1) and (b2), frontier distribution increases the UAV's exploration efficiency by 24\%. This is because hard-to-reach narrow passages  close to the floor are assigned to the UGV; the UAV is guided to regions with high information gain.

\begin{figure}
    \centering
    \includegraphics[width=\linewidth]{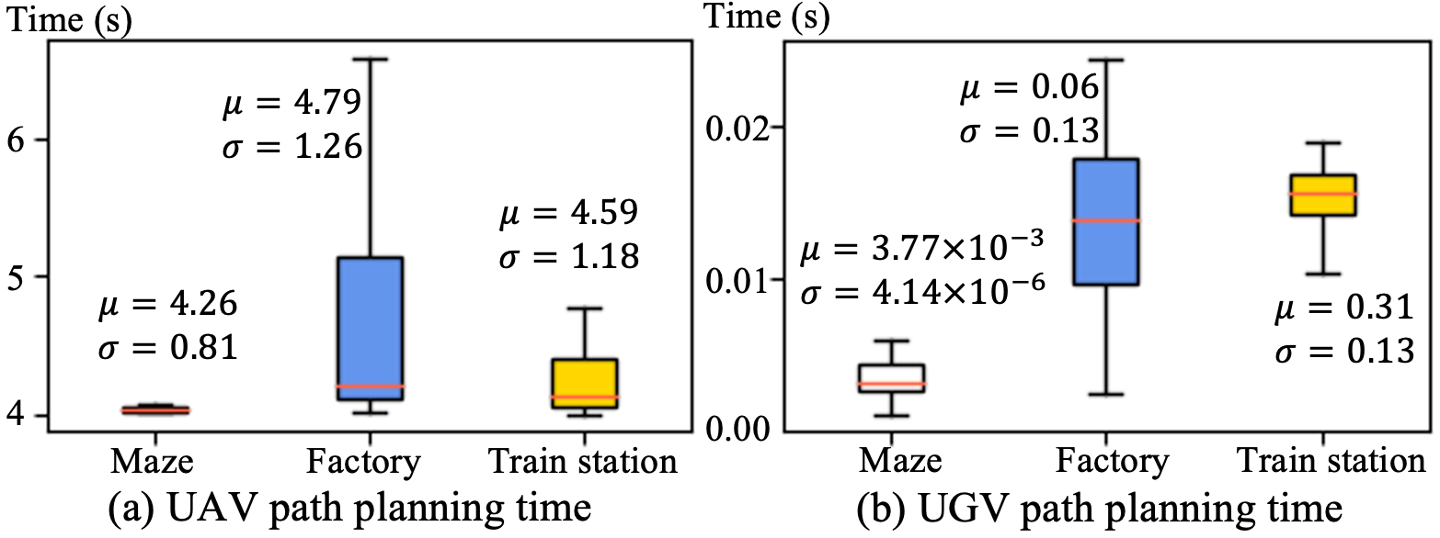}
    \caption{Path planning time for the (\textbf{a}) UAV, and the \textbf{UGV} in each step for exploring the maze, the factory, and the train station.}
    \label{fig:planning_time}
    \vspace{-0.3cm}
\end{figure}

Although we optimize paths' information gain with gradient descent, our algorithm can still run online. The proposed algorithm runs in real-time with an average total planning time in each step less than $5$s in all three environments, as shown in Fig. \ref{fig:planning_time}. The total UAV path planning time is 2.6min, 5.6min, and 10.4min, 
while the total UGV path planning time is 0.14s, 4.0s, and 18.6s using 69, 74, and 137 steps to explore the maze, the factory, and the train station, respectively.

\begin{figure}
    \centering
    \includegraphics[width=\linewidth]{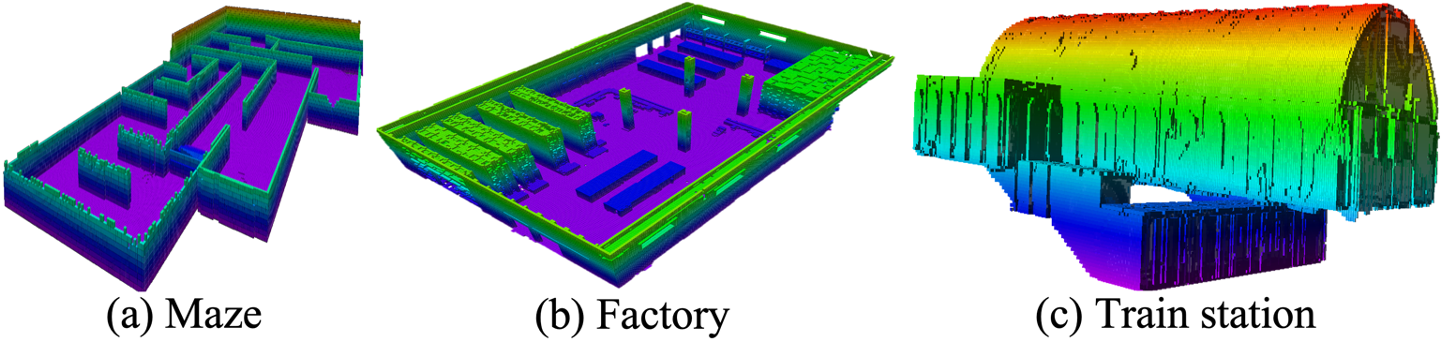}
    \caption{Visualization of explored environments: the octrees are obtained by simulating the planned paths for a corporate UAV-UGV combination using the proposed method in Airsim.}
    \label{fig:reconstruct}
    \vspace{-0.5cm}
\end{figure}
The reconstructed occupancy grid maps of the environments shown in Fig. \ref{fig:reconstruct} cover 97.2\% of the maze, 97.5\% of the factory, 96.0\% of the train station. Completely coverage of these environments is impossible because of occluded regions and agents' collision geometry. 

\vspace{0.2cm}
\section{Conclusion}
\noindent In this paper, we proposed frontier distribution and Monte-Carlo view quality rendering algorithm for a combination of a UGV equipped with a 3D lidar and a UAV equipped with an RGB-D sensor to explore unknown 3D indoor environments. Since these two processes are conducted incrementally as the explored region grows, computing the goal is real-time even in large complex environments. The global path from the current configuration to the goal is optimized using the frontier-based automatic-differentiable measure to improve information gain at each iteration and simultaneously reduce the total number of exploration iterations. The effectiveness of our proposed algorithm is verified with three simulated environments.

Current CPU-based goal selection with Monte-Carlo view quality rendering technique is not suitable for planning paths for a UAV. This is because a UAV's travel corridor has a much larger volume than the UGV's travel corridor. With limited ray samples from each frontier, the number of rays received by a viewpoint fails to represent its information gain. If the number of samples increases, a GPU-based ray casting subroutine is necessary to achieve online path planning. In the future, we will use multi-thread and shared-memory parallel programming to improve the performance of the UAV's goal selection by casting rays from all frontier voxels to the UAV's travel corridor.


Moreover, the frontier distribution and Monte-Carlo view quality rendering technique can also be applied to other robots equipped with different onboard sensors, such as automatic underwater vehicles equipped with a sonar sensor.





\clearpage
\bibliographystyle{IEEEtran}
\bibliography{main}

\end{document}